\documentclass{IOS-Book-Article}

\usepackage{mathptmx}
\usepackage{hyperref}
\usepackage{color}
\usepackage{mdframed}
\usepackage{graphicx}
\usepackage{float}
\usepackage{multirow}
\usepackage{tabularx}
\usepackage[table]{xcolor}
\usepackage{colortbl}
\definecolor{LightGray}{gray}{0.9}
\usepackage{amsmath}
\usepackage{rotating} 
\usepackage{makecell}

\begin{document}
\begin{frontmatter}              

\title{From Text to Structure: Using \\ Large Language Models to Support the Development of Legal Expert Systems}
\runningtitle{From Text to Structure}

\author[A]{\fnms{Samyar} \snm{Janatian}
\thanks{Corresponding Author: Samyar Janatian, samyar.jan10@gmail.com}},
\author[A]{\fnms{Hannes} \snm{Westermann}}, 
\author[A]{\fnms{Jinzhe} \snm{Tan}}, 
\author[B]{\fnms{Jaromir} \snm{Savelka}} and 
\author[A]{\fnms{Karim} \snm{Benyekhlef}}

\runningauthor{Janatian et al.}
\address[A]{Cyberjustice Laboratory, Universit\'e de Montr\'eal, Canada}
\address[B]{School of Computer Science, Carnegie Mellon University, USA}

\begin{abstract}
Encoding legislative text in a formal representation is an important prerequisite to different tasks in the field of AI \& Law. For example, rule-based expert systems focused on legislation can support laypeople in understanding how legislation applies to them and provide them with helpful context and information. However, the process of analyzing legislation and other sources to encode it in the desired formal representation can be time-consuming and represents a bottleneck in the development of such systems.  Here, we investigate to what degree large language models (LLMs), such as GPT-4, are able to automatically extract structured representations from legislation.  We use LLMs to create pathways from legislation, according to the JusticeBot methodology for legal decision support systems, evaluate the pathways and compare them to manually created pathways. The results are promising, with 60\% of generated pathways being rated as equivalent or better than manually created ones in a blind comparison. The approach suggests a promising path to leverage the capabilities of LLMs to ease the costly development of systems based on symbolic approaches that are transparent and explainable.
\end{abstract}

\end{frontmatter}

\thispagestyle{empty}
\pagestyle{empty}

\section{Introduction}

Legal rules often contain criteria (requirements) and conclusions (consequences) that follow should the criteria be fulfilled. For example, a judge may need to assess the criterion of whether a rent payment is a certain number of weeks late, in order to decide on the conclusion that the lease can be terminated. Sometimes, multiple cumulative criteria have to be fulfilled for a certain consequence to be enacted, or several alternative criteria could be sufficient to arrive at a legal conclusion. This structure thus informs individuals of how they should behave to fall or not fall under certain rules, lawyers of which criteria they need to establish, and judges on how they should decide on cases. In order to support these individuals in these tasks, the rules can be encoded in a formal representation to create legal expert systems using rule-based reasoning (see e.g. \cite{westermann2023justicebot,paquin1991loge,susskind1987expert}).

Understanding and encoding a legal rule is not an easy task, and may require legal training and considerable time investment. Thus, it can represent a bottle-neck in the creation of legal decision support tools (compare e.g. \cite{paquin1991loge, susskind1987expert}). In this work, we investigate whether a large language model (LLM), such as OpenaAI's GPT-4, is able to accomplish the task of analyzing legal rules and extracting a formal representation. Thus, we use GPT-4 to process a legislative text and output a structured representation. Here, we follow the approach suggested in \cite{westermann2023justicebot}, which comprises legal criteria and legal conclusion that are linked together in a pathway that shows the criteria that need to be met to arrive at given conclusions. While this task may seem complex, LLMs have recently shown strong performance for many different tasks, including in the legal domain.

LLMs being able to perform the task of extracting pathways from legislation could have important implications for the field of AI \& Law. It would allow us to leverage the power of large language models to support the creation of transparent and explainable legal expert systems using rule-based reasoning. The LLM could generate a draft pathway which can then be verified and completed by a legal expert. This could, e.g., allow for more efficient creation of legal decision support tools such as the JusticeBot \cite{westermann2023justicebot}, leading to more such tools being created and thus an increased impact on access to justice.

To investigate the capabilities of LLMs (GPT-4) in extracting structured representations from legislation, we analyze the following research question:

\begin{enumerate}
    \item RQ1 - To what degree can the LLM extract pathways from real-world legislation to be utilized in a decision support system?
    \item RQ2 - How do the pathways generated by an LLM compare to manually created pathways?
\end{enumerate}






\section{Related Work}
\label{sec:related_work}

In this research, we investigate the use of LLMs to extract structure from legislation, and encode the structure in a rule-based representation. Such rule-based representations have a long history in the field of AI \& Law. For example, \cite{allen1977normalized} presented a way to logically encode legislation to make it more syntactically clear and easier to understand. \cite{sergot1986british} implemented the British Nationality Act in Prolog. \cite{walker2006default} suggested modelling legislation in the form of an implication tree, which was used to model, e.g., Board of Veterans appeal cases \cite{walker2017semantic} and landlord-tenant disputes \cite{westermann2019using}. \cite{satoh2010proleg} introduced PROLEG, which can be used to model possible arguments by parties on either side in a rule-based manner. Here, we use the representation introduced in \cite{westermann2023justicebot}, which comprises legal criteria and legal conclusion that are linked together in a pathway.


There have further been attempts to automatically extract such structures from legislative texts. \cite{van2004case} uses a linguistic approach to detect language constructs (based on noun and verb phrases) that can then be converted to UML/OCL. \cite{nakamura2008towards} described a framework to encode legislative sentences into a logical formulation based on morphology. \cite{wyner2011rule} extracts deontic rules from legislation using a rule-based approach, based on annotated legal texts. \cite{gaur2015translating} built a system to translate simple sentences to various logical representations.  \cite{dragoni2016combining} used a combination of approaches, including tree parsers, to parse rules form legislation. Attempts to make legal rules computable have further been made, e.g., for self-driving vehicle traffic rule compliance checking \cite{bhuiyan2022traffic,sasdelli2022survey}. Beyond the legal domain, language models have been used to some degree to build causal graphs, e.g., in the medical domain to confirm causal pathways \cite{long2023large}. Here, we use LLMs to extract a pathway from plain-text, real-world legislation and export them in a format suitable for integration in a JusticeBot tool. To our knowledge, this is the first attempt that uses LLMs to directly convert a text to a structured legal representation.

We use LLMs to extract structure from legislation. Models such as OpenAI's GPT-4 \cite{openai2023gpt4} have shown to be capable in handling many diverse tasks. They have also been used in the legal domain to, e.g., pass the bar exam \cite{katz2023gpt} and perform statutory reasoning \cite{blair2023can, nguyen2023blackbox}. Such models have further been investigated for the annotation of legal documents \cite{savelka2023can}, to explain the meaning of statutory terms \cite{savelka2023explaining}, to mediate disputes \cite{westermann2023llmediator}, and to provide legal information \cite{tan2023chatgpt}. Here, we use LLMs to automatically extract a structured representation from legislative text.

The main focus of this paper is to support a legal expert in performing the tedious work of extracting a formal representation from legislation. Such efforts are in line with a large body of previous work in AI \& Law to, e.g., support legal experts in efficient annotation of texts \cite{branting2021scalable,westermann2020sentence}, to discover boolean search rules \cite{westermann2019computer} 
or to create legal case briefs \cite{westermann2022toward}. In this work, we aim to understand whether our approach can generate plausible drafts of legal pathways, that can serve as a basis for, e.g., a legal decision support tool.






\section{Proposed Framework}
\begin{figure}[h]
\centering
\includegraphics[width=1\textwidth]{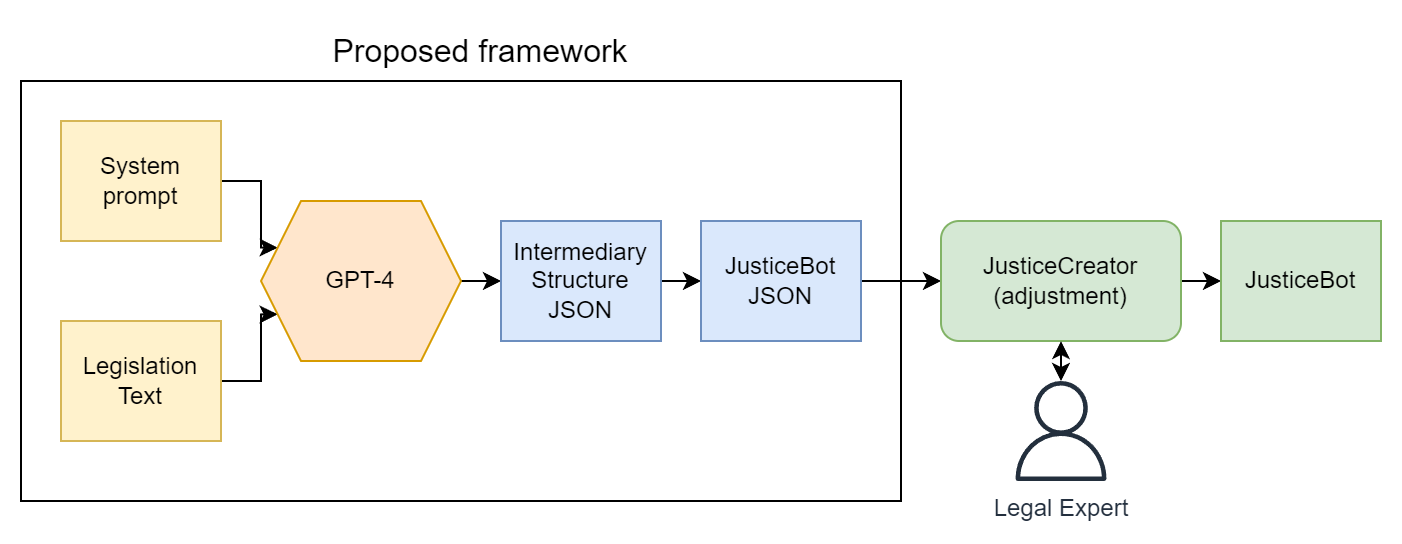}
\caption{The steps of the proposed framework, from legislation text to JusticeBot expert system.}
\label{fig:proposed_framework}
\end{figure}

We propose and evaluate a framework that takes a piece of legislation (a single article or paragraph) and extracts a pathway of the rules contained therein, using an LLM. The pathway can be imported into the ``JusticeCreator'' \cite{westermann2023justicebot}, used to create JusticeBot tools,  and further edited by legal experts, to verify the accuracy and improve the pathway if needed and add further content. We have built a tool called the JusticeCreator Automatic Pathway Generator (JCAPG) to make this functionality accessible to the legal experts. Figure \ref{fig:proposed_framework} gives an overview of the different steps involved in the framework.

\subsection{Analysis of legislation}
In order to extract structured representations from legislation, we provide GPT-4 with the text of a piece of legislation as a user message. GPT-4 is also given a system message (used to instruct the model), with instructions to extract the requirements and legal conclusions from the legislation, and create the links between these elements (see \ref{sec:pathway_creation}). 

We submit the system message and user message to the model as a prompt. GPT-4 returns a JSON response of the logic blocks and their connections, which is then converted to the JSON format that is compatible with the JusticeCreator (see figure \ref{fig:proposed_framework}).

The final JSON file can be exported from the JCAPG, and imported into the JusticeCreator, where legal experts can verify and adjust the pathway, e.g., by correcting logical errors if any. The exported file can thus serve as a starting point for the creation of a JusticeBot tool covering a new legal area, potentially making the process more efficient.

\section{Experimental Design}
In order to evaluate the capability of LLMs to extract pathway structure from legislation, we devised an experiment to select a set of legal articles, analyze aspects of the generated pathways and compare the generated pathways to manually created ones. 

\subsection{Selection of Articles}

In order to test the ability of LLMs to automatically extract structured representations from articles, we specifically selected stand-alone articles or paragraphs containing descriptions of criteria and conclusions, which did not depend on nor reference other articles.
We screened the articles of the Civil Code of Quebec to find 40 articles that fit the criteria, and selected these for our experiments. The articles we selected range from very simple (one requirement linked to one outcome), to moderately complex, involving several connected criteria and legal conclusions. For this initial experiment, we excluded overly complicated articles that would pose considerable challenge even to domain experts. Table \ref{fig:article_stats} contains statistics about the articles, grouped by the complexity of the article assigned by the annotators.


\begin{table}[h]
    \centering
    \caption{Average Data about the 40 CCQ Articles grouped by their respective Difficulty}
    \begin{tabular}{|c|c|c|c|c|c|c|c|}
        \hline
        \textbf{Difficulty} & \makecell{\textbf{Number of } \\ \textbf{Articles (\(N\))}} & \makecell{\textbf{Character} \\ \textbf{Count}} & \makecell{\textbf{Token} \\ \textbf{Count}} & \makecell{\textbf{Manual} \\ \textbf{Time}} & \makecell{\textbf{Automatic} \\ \textbf{Time}} & \makecell{\textbf{Manual} \\ \textbf{Nodes}} & \makecell{\textbf{Automatic} \\ \textbf{Nodes}} \\
        \hline
        Easy         & 18 & 202.17 & 40.83 & 3.19 min. & 19.18 sec. & 4.06 & 4.61 \\
        \hline
        Normal       & 14 & 256.50 & 52.00 & 5.66 min. & 22.84 sec. & 5.29 & 5.43 \\
        \hline
        Hard         &  8 & 371.38 & 74.38 & 7.32 min. & 27.54 sec. & 6.13 & 6.00 \\
        \hline
    \end{tabular}
    \label{fig:article_stats}
\end{table}


\subsection{Pathway creation}
\label{sec:pathway_creation}
We used the selected articles to create pathways, both manually and with the JCAPG. The methodology used is largely similar to that described in \cite{westermann2023justicebot}, of encoding legal criteria as ``questions'' blocks, and linking their presence or absence to further legal criteria or legal outcomes. However, instead of creating interviews that speak to a layperson in simplified language, we focus on third-person formulations of the legal criteria.


Four annotators (co-authors of this paper) created \textbf{manual representations} of the pathways. All of the annotators have some legal education, and 3 had previous experience in using the JusticeCreator. 
The annotators were asked to rate the ``complexity'' of the articles after pathway creation, capturing the structural complexity and ambiguity of the article, and to note how much time creating the pathway took (see table \ref{fig:article_stats}). 

Next, we used the JCAPG method to convert these articles to structured pathways \textbf{automatically}. We used the GPT-4 model, through the \texttt{openai} Python library\footnote{GitHub: OpenAI Python Library. Available at: \url{https://github.com/openai/openai-python} [Accessed 2023-09-17]}. We used default model parameters, except the \texttt{temperature} value, which we set to 0 to decrease the risk of the model hallucinating and making the output more reproducible.

We iteratively optimized the prompt using 4 articles (not part of the test data).\footnote{Prompt and code to run the analysis available at: \url{https://github.com/samyarj/JCAPG-JURIX2023}} For example, we noticed that the model would occasionally generate recursive pathways which would crash our JSON parser module, and included instructions to avoid this. We also instructed the model to avoid the logical error of denying the antecedent, with partial success (see section \ref{sec:discussion_rq1}). Finally, we ran the JCAPG on the 40 selected articles to create the pathways for the experiment.



\begin{figure}[h]
\centering
\includegraphics[width=1.0\textwidth]{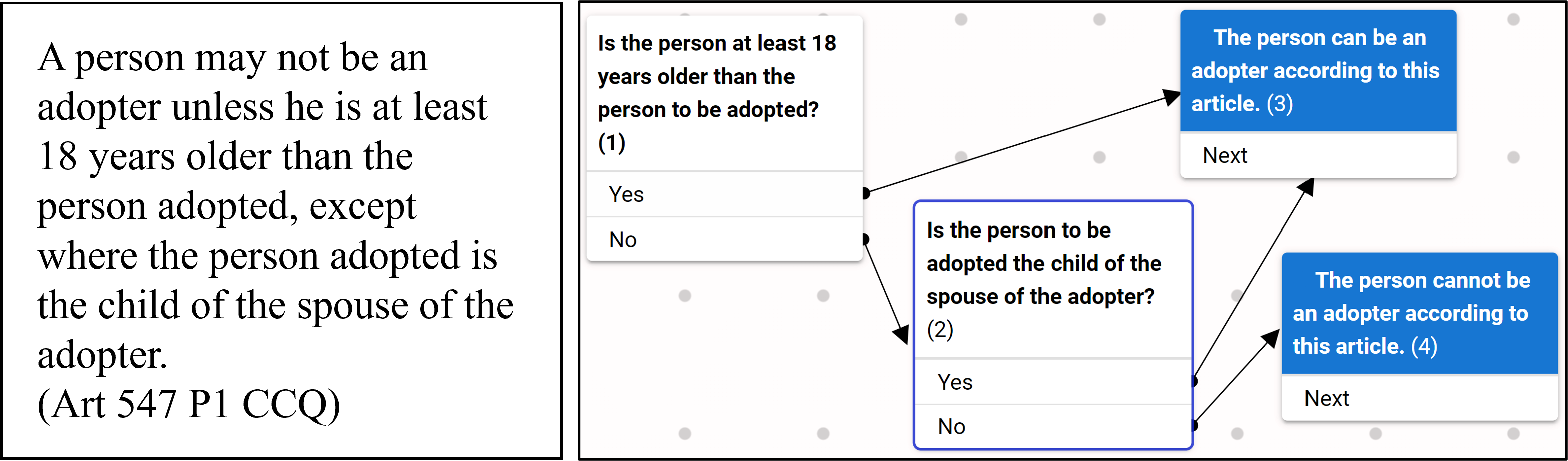}
\caption{A legal article and the corresponding automatically generated pathway.}
\label{fig:Pathway display}
\end{figure}


\subsection{Evaluation}
\textbf{E1 - Evaluation of automatically generated pathways}
In the first experiment, each annotator was asked to rate the automatically generated pathways and compare it to their own manual pathway, in light of the legislation. They were asked to answer questions that, e.g., analyzed the textual and structural accuracy of the article, and its usefulness as a first draft of a pathway (see table \ref{tab:evaluation_metrics}).

\noindent\textbf{E2 - Blind test comparison between human and automatically generated pathways}
We also performed a blind test, to compare the human generated pathways to the automatic ones. Each annotator was given two pathways for the same article, one created automatically and one manually, without knowing which was which, and asked to indicate which of the pathways they preferred in different aspects (see table \ref{fig:blind-test}). 



\section{Results}

\subsection{E1 - Evaluation of automatic pathways}

Table \ref{tab:evaluation_metrics} shows the results of the evaluation of the generated pathways. As we can see, the textual accuracy is very high at 92.5\%. Logical accuracy and lack of hallucination were slightly weaker, with 72.5\% and 87.5\% respectively. Even though only 20\% of the generated articles matched the manually created pathways, 67.5\% of generated pathways were rated to be of equal logical quality as the manually created pathways. While 40\% of the automatically generated pathways were rated as directly usable, an additional 50\% were deemed as needing only slight adjustments.


\begin{table}[h]
    \centering
    \caption{Evaluation Metrics for the 40 Automatically Generated Pathways}
    \begin{tabularx}{\textwidth}{|X|c|c|c|}
        \hline
        \textbf{Evaluation Criteria} & \textbf{Result} & \textbf{\(N\)} & \textbf{\%} \\
        \hline
        \multirow{2}{5cm}{Textual Accuracy - Does the textual content of pathway match the content in the law?}   & \cellcolor{green!40} Yes  & 37 & 92.5 \\
                                            & \cellcolor{red!40} No   & 3 & 7.5 \\
        \hline
        \multirow{2}{5cm}{Completeness - Are all logical elements from the law contained in the pathway?}       & \cellcolor{green!40} Yes   & 29 & 72.5 \\
                                            & \cellcolor{red!40} No  & 11 & 27.5 \\
        \hline
        \multirow{2}{5cm}{No Hallucination - The model did not invent criteria or conclusions.}   & \cellcolor{green!40} Yes   & 35 & 87.5 \\
                                            & \cellcolor{red!40} No  & 5 & 12.5  \\
        \hline
        \multirow{2}{5cm}{Matching - Did the pathway logic perfectly match the manual pathway?} & \cellcolor{green!40} Yes   & 8 & 20.0 \\
                                             & \cellcolor{red!40} No  & 32 & 80.0 \\
        \hline
        
        \multirow{4}{5cm}{Overall Rating - How well suited would the generated pathway be to serve as the basis of a JusticeBot tool?} 
        & \cellcolor{blue!40} Correct                  & 16 & 40.0  \\
        & \cellcolor{green!40} Slight Adjustment Necessary   & 20 & 50.0 \\
        & \cellcolor{yellow!40} Starting Point  & 4 & 10.0  \\
        & \cellcolor{red!40} Useless  & 0 & 0.0  \\                                     
        \hline
    \end{tabularx}
    \label{tab:evaluation_metrics}
\end{table}

Table \ref{tab:combined_table} (left) shows how humans rated the machine generated articles, when split by the rated difficulty level of the article. For the easy articles, a majority was rated as correct, whereas this number was lower for medium and hard articles.

\begin{table}[h]
    \centering
    \caption{Left: Overall ratings of generated pathways. Right: Results of blind study overall comparison between manual and generated articles. Both Grouped by The Difficulty of The Legislation Articles}
    \small 
    \begin{minipage}{0.48\textwidth}
        \centering
        \setlength\tabcolsep{4pt}
        \begin{tabular}{|c|c|c|c|}
            \hline
            \textbf{Diffic.} & \textbf{Overall Rating} & \textbf{\(N\)} & \textbf{\%} \\
            \hline
            Easy  & \cellcolor{yellow!40}Starting point   & \cellcolor{yellow!40}1 & \cellcolor{yellow!40}5.6 \\
               & \cellcolor{green!40} Adjustment Necessary  & \cellcolor{green!40}6 & \cellcolor{green!40}33.3 \\
               & \cellcolor{blue!40} Correct  & \cellcolor{blue!40}11 & \cellcolor{blue!40}61.1 \\
            \hline
            Normal  & \cellcolor{yellow!40} Starting point  & \cellcolor{yellow!40}1 & \cellcolor{yellow!40}7.1 \\
               & \cellcolor{green!40} Adjustment Necessary  & \cellcolor{green!40}10 & \cellcolor{green!40}71.4 \\
               & \cellcolor{blue!40} Correct  &  \cellcolor{blue!40}3 &  \cellcolor{blue!40}21.4 \\
            \hline
            Hard  & \cellcolor{yellow!40} Starting point  & \cellcolor{yellow!40}2 & \cellcolor{yellow!40}25.0 \\
               & \cellcolor{green!40} Adjustment Necessary & \cellcolor{green!40}4 & \cellcolor{green!40}50.0 \\
               & \cellcolor{blue!40} Correct  & \cellcolor{blue!40}2 & \cellcolor{blue!40}25.0 \\
            \hline
        \end{tabular}
    \end{minipage}
    \begin{minipage}{0.48\textwidth} 
        \centering
        \setlength\tabcolsep{4pt} 
        \begin{tabular}{|c|c|c|c|}
            \hline
            \textbf{Diffic.} & \textbf{Preferred Pathway} & \textbf{\(N\)} & \textbf{\%} \\
            \hline
            Easy  & \cellcolor{green!40}Automatic & \cellcolor{green!40}9 & \cellcolor{green!40}50.0 \\
               & \cellcolor{yellow!40}Equivalent  & \cellcolor{yellow!40}5 & \cellcolor{yellow!40}27.8 \\
               & \cellcolor{red!40}Human    & \cellcolor{red!40}4 & \cellcolor{red!40}22.2 \\
            \hline
            Normal  & \cellcolor{green!40}Automatic & \cellcolor{green!40}3 & \cellcolor{green!40}21.4 \\
               & \cellcolor{yellow!40}Equivalent  & \cellcolor{yellow!40}4 & \cellcolor{yellow!40}28.6 \\
               & \cellcolor{red!40}Human    & \cellcolor{red!40}7 & \cellcolor{red!40}50.0 \\
            \hline
            Hard  & \cellcolor{green!40}Automatic & \cellcolor{green!40}3 & \cellcolor{green!40}37.5 \\
               & \cellcolor{yellow!40}Equivalent  & \cellcolor{yellow!40}0 & \cellcolor{yellow!40}0 \\
               & \cellcolor{red!40}Human   & \cellcolor{red!40}5 & \cellcolor{red!40}62.5 \\
            \hline
        \end{tabular}
    \end{minipage}%
    \label{tab:combined_table}
\end{table}

\subsection{E2 - Evaluation of blind test experiment}
\begin{table}[h]
    \setlength{\tabcolsep}{3pt}
    \centering
    \caption{Summary of The Blind Test}
    \begin{tabular}{|l|c|c|c|}
        \hline
        \textbf{Question asked to the Evaluators during Blind Test} & 
        \makecell{\textbf{Automatic} \\ \textbf{Pathway}} & \textbf{Equivalent} & \makecell{\textbf{Manual} \\ \textbf{Pathway}} \\
        \hline
        Which Pathway is better overall? & 15 (37.5\%) & 9 (22.5\%) & 16 (40.0\%) \\
        \hline
        Which Pathway Better Reflects the Content of the Law? & 14 (35.0\%) & 16 (40.0\%) & 10 (25.0\%) \\
        \hline
        Which Pathway Better Reflects the Logical Structure of the Law? & 14 (35.0\%) & 10 (25.0\%) & 16 (40.0\%) \\
        \hline
    \end{tabular}
    \label{fig:blind-test}
\end{table}

Table \ref{fig:blind-test} shows the result of the blind test. Overall, the preference was evenly split between the generated and the manual pathways. Table \ref{tab:combined_table} (right) shows whether the annotators overall preferred the automatic or manually created pathways, based on the difficulty level of the article. While for the easy articles the automatic pathway was rated as better or equivalent in 77.8\% of cases, for the hard articles this percentage dropped to 37.5\%.

\section{Discussion}


\subsection{RQ1 - To what degree can the LLM extract pathways from real-world legislation to be utilized in a decision support system?}
\label{sec:discussion_rq1}
To evaluate to what degree LLMs can extract pathways from legislation, we assessed pathways generated from 40 articles of the Civil Code of Quebec. The results of the evaluation are presented in table \ref{tab:evaluation_metrics}. The \textit{textual accuracy} of the generated pathways is high, with 92.5\% of them being rated as textually accurate. This shows that our prompt asking the model to stick to the legislative text and providing the legislative text in the prompt seems like a viable method to constrain the textual output to the targeted legislative text. The \textit{Completeness} of the generated pathways is somewhat lower, with 72.5\% of the generated articles being rated as containing all of the criteria and conclusions from the law. Likewise, the model in 87.5\% of cases sticks to the elements contained in the law. Overall, the generated pathways were rated as ``Correct'' in 40\% of the cases, and requiring only a slight adjustment in 50\% of the cases, which is a promising result. 


An interesting aspect of the results is that the articles that were rated as difficult by the annotators also generally lead to lower scores for the generated pathways, as can be seen in table \ref{tab:combined_table}. For the easy articles, 61.1\% of the generated pathways were rated as correct, while for the normal and hard articles, only 21.4\% and 25\% were rated as correct. Some of this variance comes down to the \textit{ambiguity} of the legal articles, which was also frequently highlighted in comments by the annotators. This mirrors prior discussion in the field (see, e.g. \cite{ashley2017artificial}), showing that statutory language is not always clear enough to yield an unambiguous interpretation. Doctrine, court cases and domain expertise may be needed to elicit such an interpretation in these cases. When the article was ambiguous, the model would occasionally rely on (reasonable) assumptions not contained in the article.

Even if the article was clear, we noticed that there was often no single ``right'' way to split criteria and conclusions into different elements. While some of these may be preferable, the other ones are not technically wrong. Sometimes, the model would, e.g., introduce a criterion in the final conclusion block, instead of as a question. Likewise, if an article contained multiple conclusions, there was no ``correct'' way to create the pathway, since we excluded intermediary conclusions for this experiment. This type of difference explains the low structural correspondence between manually and automatically created pathways, but does not necessarily impact the usefulness of the generated pathway.

Another error the model made occasionally was ``denying the antecedent'', i.e., incorrectly inferring conclusions not  in the article. There were also more simple errors of the model, such as misunderstanding the structure of an article, or even generating a pathway with multiple starting points or disconnected blocks. However, these were rare.

\subsection{RQ2 - How do the pathways generated by an LLM compare to manually created representations?}
Next, we conducted a blind test to see how the generated articles compared to manually created articles. The results are presented in table \ref{fig:blind-test}. Again, the model shows strong performance for the textual content of the law, where only 25\% prefer the manually created pathway. For the logic and overall ratings, the results are slightly weaker, with 40\% of the annotators preferring the manually created pathways in these aspects. However, even in those categories, 60\% of the annotators either preferred the automatic pathway or saw them as equivalent, which is a positive result.

The results are affected by the difficulty of the article. For simple articles, the automatically created pathway was somewhat preferred over the manually created one, with only 22\% of annotators saying they prefer the latter. For the normal articles, this rose to 50\%, and for the hard articles, 62.5\% preferred the manually created pathway. 
Thus, it seems like the ability of the model to create a pathway from legislation compares favourably to human annotators, especially for the case of easy articles. 
For the more difficult articles, the model sometimes produces an incorrect output, e.g., missing important elements or basing its analysis on assumptions that are not part of the text.

\subsection{Use as augmented intelligence}
Next, let us discuss the potential of the tool as ``augmented intelligence'' (or ``hybrid intelligence''), that suggests a pathway to the expert annotator, who can then evaluate it and adjust it if necessary. 
A first indication can be gleaned from the final question we asked the experts in E1, of whether a pathway would need to be adjusted to serve as the basis for a JusticeBot decision support tool. As can be seen in table \ref{tab:evaluation_metrics}, the annotators found that 90\% of the generated pathways were either correct, or would only require a small adjustment. None were found to be useless. 
Even for the hard articles, the generated pathways were seen as being correct or useful after an adjustment in 75\% of cases. The majority of generated pathways were thus seen as being useful, which is promising.

Even more interestingly, when looking at table \ref{fig:blind-test}, we can see that the pathways generated by the JCAPG were rated as better than the pathways generated by humans in 37.5\% of cases. It seems like in some instances, the model can capture nuances or logical particularities that the humans missed. This phenomenon is also captured in the comments left by the annotators, where in five instances human annotators discovered logical errors in their own reasoning after reading the automatically generated pathway.


These results seem to indicate that the JCAPG, used in conjunction with a human expert, has the potential to support the annotator with a strong draft, resulting in more efficient annotation, and even leading to more logically correct pathways. Thus, LLMs can potentially be used to support a human in more efficiently creating predictable and safe legal expert systems, which could have beneficial impacts on, e.g., access to justice.

\subsection{Limitations}
It is important to consider limitations of the study, which is intended to serve as an initial validation of the approach. First, the paper is based on a selection of articles from the Civil Code of Quebec. This legislation stems from the civil code tradition, which focuses on drafting articles that are easy to understand for laypeople \cite{bergel1987principal}. It is possible that the results would not translate to jurisdictions that do not have this aim. Second, we deliberately selected articles that were both standalone, and relatively straightforward, even though some articles were of moderate complexity. The results may not replicate to very complex articles, or articles that can only be read in conjunction with other articles (see future work). Third, our first experiment is not conducted in a blind fashion, with relatively few annotators, which means that biases may have entered into the analysis. Fourth, even though the approach seems promising as a way to support annotators in creating pathways, it is important to note that the structuring of pathways is only a part of creating a JusticeBot legal decision support tool as described in \cite{westermann2023justicebot}. Additionally, the work also requires the simplification of the content, the drafting of layperson explanations as to the individual criteria, and the addition of case law summaries to the individual question blocks (see \cite{salaun2022conditional} regarding automatic summarization).

\section{Conclusion and Future Work}
We investigated the capability of LLMs to extract a pathway capturing legal criteria and conclusions from legislative texts. The human evaluators rated 40\% of the automatically generated pathways as directly usable (i.e., correct), and additional 50\% as needing only slight adjustment. In the blind test we conducted, 60\% of the generated pathways were rated as equivalent or better than manually created ones. These results are promising, especially to support legal experts by providing a draft as a basis for a legal expert system. In \textit{future work}, we aim to extend the experiment to involve more annotators and more complex legal articles (in order to investigate the generalizability of the approach), to involve multiple connected legal articles instead of a single article, to potentially integrate case law and/or doctrine to resolve ambiguities, to study the quality and efficiency of using the generated pathways as drafts for annotators and to integrate LLMs in other ways, e.g., by mapping case facts to the generated pathways (compare \cite{nguyen2023logic,westermann2023bridging}).

\noindent\rule{0pt}{4ex}\textbf{Acknowledgements} We acknowledge the generous support from the Cyberjustice Laboratory, LexUM Chair, and Autonomy through Cyberjustice Technologies project.



\bibliography{biblio}

\end{document}